\documentclass{article} % For LaTeX2e
\usepackage{iclr2027_conference,times}

\usepackage{amsmath,amsfonts,bm}

\def\eqref#1{equation~\ref{#1}}
\def\1{\bm{1}}

\DeclareMathAlphabet{\mathsfit}{\encodingdefault}{\sfdefault}{m}{sl}
\SetMathAlphabet{\mathsfit}{bold}{\encodingdefault}{\sfdefault}{bx}{n}

\definecolor{revisionpink}{HTML}{C73E63}
\newcommand{\rev}[1]{\textcolor{black}{#1}}

\usepackage{hyperref}
\usepackage{url}
\usepackage{wrapfig}
\usepackage{multirow}
\usepackage{graphicx}
\usepackage{booktabs}
\usepackage[table]{xcolor}
\usepackage{bbm}
\usepackage{amsmath}
\usepackage{float}

\title{
Spec-CoFrGeNet: Reparametrizing Continued Fraction Networks for Stable and Accurate Training}

\author{%
  Soumyadeep Pal\\
  Michigan State University \\
  \texttt{palsoum1@msu.edu} \\
  \And
  Amit Dhurandhar\\
  IBM Research \\
  \texttt{adhuran@us.ibm.com} \\
  \And
  Vijil Chenthamarakshan \\
  IBM Research \\
  \texttt{ecvijil@us.ibm.com} \\
  \AND
  Dennis Wei \\
  IBM Research \\
  \texttt{dwei@us.ibm.com} \\
 \And
  Karthikeyan Natesan Ramamurthy \\
  IBM Research \\
  \texttt{knatesa@us.ibm.com} \\
}

\newcommand{\name}{Spec-CoFrGeNet }
\newcommand{\gpt}{GPT-2 XL }
\newcommand{\granite}{Granite 4.1 }
\newcommand{\llama}{Llama-MoE }

\iclrfinalcopy % Uncomment for camera-ready version, but NOT for submission.
\begin{document}

\maketitle
\lhead{}

\begin{abstract}
Continued fractions have been shown to be a function class capable of constructing expressive yet parameter-efficient neural networks. These neural networks are not only effective for supervised learning, but also for training large language models (LLMs). However, the rational form of this function class introduces poles when the denominator approaches zero, which prior work has avoided by enforcing a small positive (non-smooth) lower bound. We show that performance and training stability can be highly sensitive to the choice of this lower bound. In this work, we introduce a novel transformation of the partial denominators, inspired by its connection to the Green's function in physics, that mitigates the above issue elegantly by constraining the spectra of the constructed tridiagonal Hamiltonian. Our approach further eliminates the need for incremental training used by previous works for stability, thus avoiding involved modifications to standard pre-training procedures. We demonstrate the effectiveness of our reparametrization by replacing FFNs with a previously proposed continued fraction architecture for both language and image generative modeling. 
Across diverse models, such as GPT-2 XL, (nano) Llama-MoE, and Granite, we show that our proposed method achieves performance comparable to the original (FFN-based) architectures while \rev{maintaining} $\mathord{\sim}30\%-50\%$ parameter savings \rev{as before}, and in most cases even outperforms the more complicated incremental training baseline. 
We further demonstrate, for the first time for continued fraction networks, that their effectiveness also carries over to image generative modeling, that utilize diffusion objectives. 
On a state-of-the-art Diffusion Transformer, we show that our approach again achieves performance comparable to the original FFN-based models with $\mathord{\sim}30\%$ parameter savings and significantly outperforms the previous incremental training approach. Code is available at \url{https://github.com/IBM/CoFrNets/}.
\end{abstract}

\section{Introduction}
\label{sec: intro}

Generative AI has been rapidly and widely adopted in the recent years, being integrated into consumer, workspace and organizational settings \citep{bick2026rapid,maslej2025artificial}. This boom in adoption is driven by advances in several generative modeling paradigms, most notably for large language models (LLM) for language generation \citep{brown2020language,ouyang2022training} and diffusion models for image generation \citep{peebles2023scalable,brooks2024video,labs2025flux}. 
An important architectural component of these networks is the feedforward network (FFN) or multilayer perceptron (MLP), which have been shown to be universal function approximators \citep{hornik1989multilayer}. 
Recent works have proposed alternatives such as continued fraction networks (CoFrNets) \citep{puri2021cofrnets,dhurandhar2026cofrgenet}, which are also universal approximators, but additionally can represent functions compactly using much fewer parameters than FFNs.
% \begin{figure}[h]
\begin{wrapfigure}{r}{.52\textwidth}
\vspace{-.1cm}
\centering
\includegraphics[width=.5\textwidth]{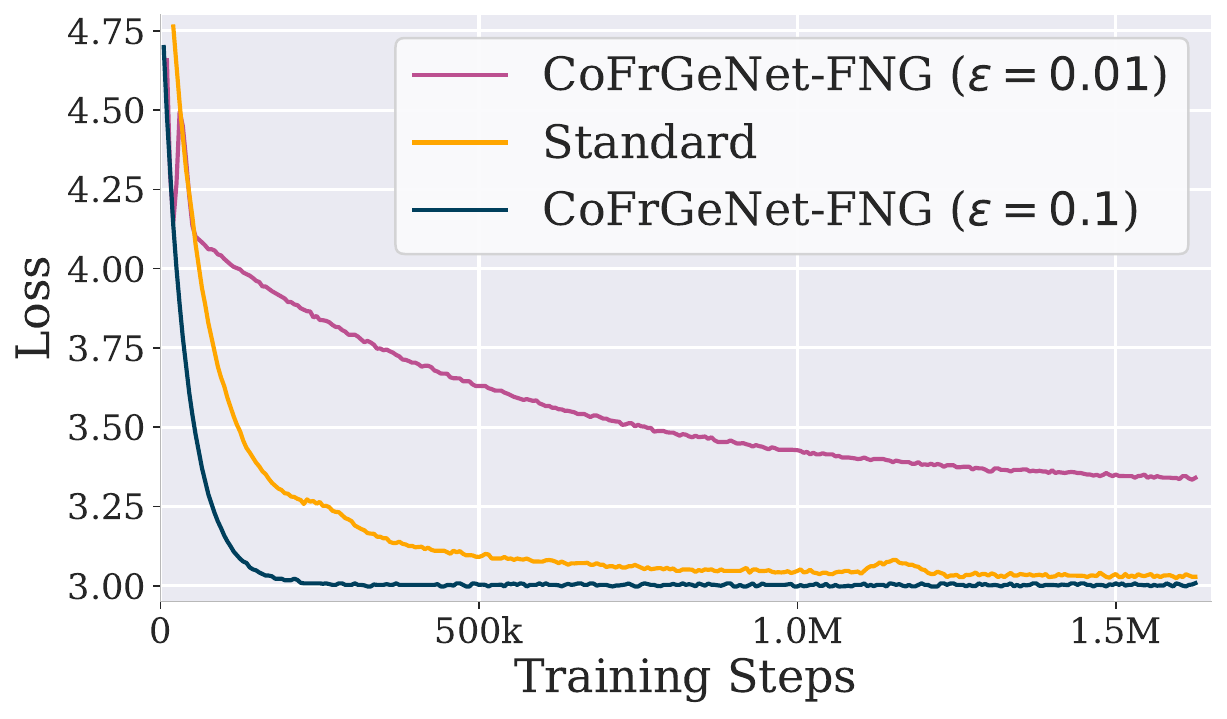}
\vspace{-.1cm}
\caption{\small{Training loss of \textbf{Llama-MoE} for the standard model and CoFrGeNet version with incremental training \citep{dhurandhar2026cofrgenet}. FNG refers to FFNs replaced with gated continued fraction ladders without any expansion (See \textbf{Sec.\,\ref{sec: background}} and \textbf{Sec.\,\ref{sec: experiment}}). Stability of CoFrGeNet is sensitive to the choice $\epsilon$, which is used to avoid poles.}
}
\label{fig:epsissue}
\vspace{-1.05cm}
\end{wrapfigure}
% \end{figure}
We now provide some background on CoFrNets necessary to motivate our formulation with more background details in Section \ref{sec: background}. 
A generalized continued fraction takes the form 
\[
a_0 + \cfrac{b_1}{a_1 + \cfrac{b_2}{a_2 + \cdots}},
\]
where the partial denominators $a_k$ and partial numerators $b_k$ may, in general, be complex-valued.

Through equivalence transformations, we can also represent this in a form where the numerators are normalized to one. This representation, referred to as the \emph{canonical form}, is given by:
\[a_0 + \cfrac{1}{a_1 + \cfrac{1}{a_2 + \cdots}}.
\]
Note that we can truncate these continued fractions to certain depths, which are called \emph{convergents} as they converge to the desired value.
In CoFrNets~\citep{puri2021cofrnets}, each of these partial denominators were parameterized using learnable affine functions of the input $a_k = w_kx$\footnote{The bias term was absorbed in $x$.}, to form a \emph{ladder}. A network is instantiated from this representation by ensembling multiple such ladders, with the reciprocal operation serving as its nonlinearity.
Considering $a := (a_1,\ldots,a_d)$, 
a ladder can then be written as
\[
f(a_0,a)
=
a_0
+
\cfrac{1}{a_1+
\cfrac{1}{\ddots+
\cfrac{1}{a_{d-1}+
\cfrac{1}{a_d}}}}
=
a_0+\tilde{f}(a),
\] where $a_0$ is the leading linear term and $\tilde{f}(a)$ denotes the non-linear component of the ladder. 
This finite continued fraction in fact admits an equivalent representation as a ratio of two polynomials, known as \emph{continuants} \citep{puri2021cofrnets}:
\begin{equation}
f(a_0,a)
=
\frac{K_{d+1}(a_0,\ldots,a_d)}
     {K_d(a_1,\ldots,a_d)}.
\label{eq:continuant_ratio}
\end{equation}
The continuants $K_k$ satisfy the recurrence
\[
\begin{aligned}
K_0 &= 1, \qquad
K_1(a_d) = a_d, \\
K_k(a_{d-k+1},\ldots,a_d)
&=
a_{d-k+1}K_{k-1}(a_{d-k+2},\ldots,a_d)
+
K_{k-2}(a_{d-k+3},\ldots,a_d).
\end{aligned}
\]
The fractional part $\tilde{f}(a)$ consequently can be expressed as $\frac{K_{d-1}(a_2,\ldots,a_d)}
     {K_d(a_1,\ldots,a_d)}$.  

One practical challenge in training continued-fraction networks is the presence of \emph{poles}. This is apparent from \eqref{eq:continuant_ratio}, where the fractional component is explicitly a rational function and therefore diverges whenever its denominator approaches zero i.e. $K_d(a_1,\ldots,a_d)=0$.
To mitigate this numerical instability, \cite{puri2021cofrnets} used the following stabilization:
\begin{equation}
K_d
\;\longrightarrow\;
\operatorname{sgn}(K_d)
\max\left(|K_d|,\epsilon\right),
\qquad
\epsilon>0.
\label{eq:eps}
\end{equation}
This clipping ensures that the magnitude of the denominator never falls below $\epsilon$, thus ensuring that the the continued fraction ladders never encounter a pole during training.

\cite{dhurandhar2026cofrgenet} used the above principles of using continued fractions for training neural networks and extended that to a generative training framework to train LLMs. 
\begin{wrapfigure}{r}{.6\textwidth}
% \vspace{-.cm}
% \begin{figure}[t]
%\vspace{-.5cm}
\centering
\includegraphics[width=.58\textwidth]{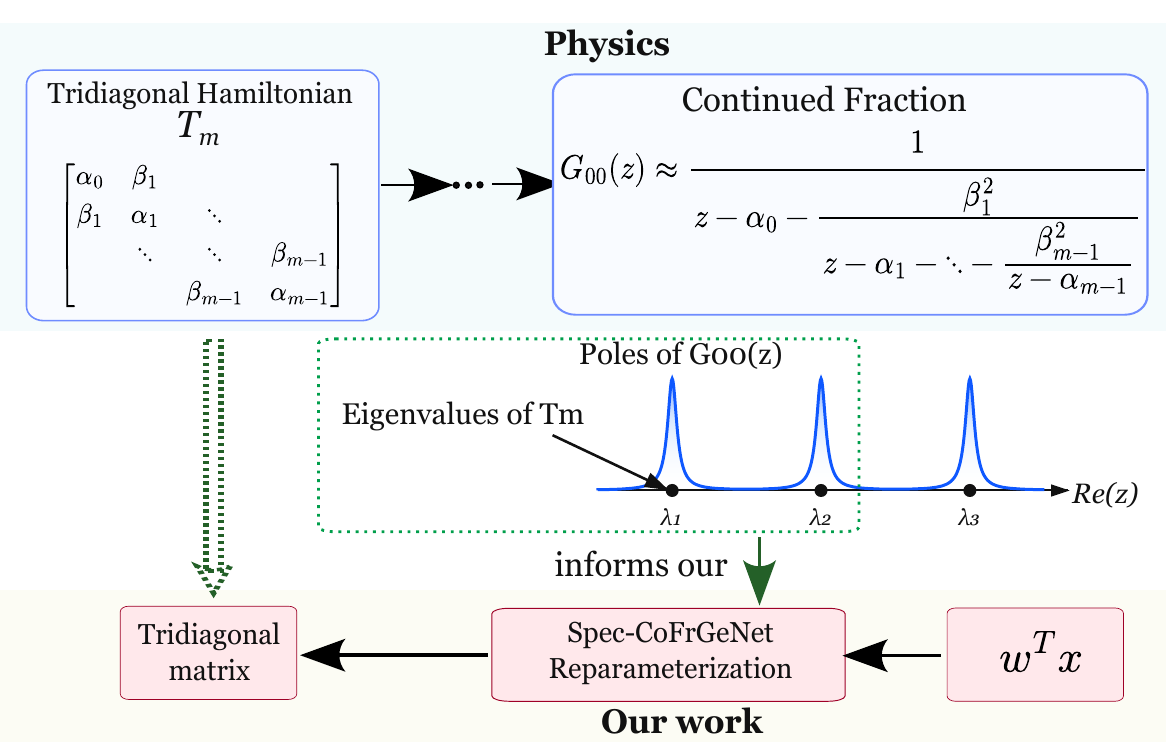}
% \vspace{-.2cm}
\caption{\small{A Hamiltonian is transformed into a tridiagonal matrix in the Lanczos construction, whose Green's function has continued fraction representation. Poles of this fraction lie at the eigenvalues of the matrix. Motivated by this connection in physics, our conceptual contribution goes the other way, where our reparameterization constrains the spectrum of the associated tridiagonal matrix.
}}
\label{fig:approach}
% \vspace{-0.6cm}
% \vspace{-1.05cm}
\end{wrapfigure}
% \end{figure}
Despite having success in doing so, it remains unclear if such networks can be adopted to generative modeling of vision using diffusion style flow-matching losses or even other architectures such as Mixture-of-Experts (MoEs) for language. 
Additionally, in our work we find that the aforementioned $\epsilon$ based stabilization has some disadvantages. As shown in Fig.\,\ref{fig:epsissue}, the convergence of the training loss relies heavily on the choice of $\epsilon$: smaller epsilon values (i.e. $\epsilon = 0.01$) lead to slower and less stable convergence. Conversely, a larger $\epsilon$ value can restrict the continuant value and decrease the range of expressible values. \emph{More importantly, we find that when using diffusion style loss for vision generative modeling, the training often gets unstable even for larger $\epsilon$ such as $0.1,~0.2$.}

In this work, we introduce \underline{Spec}tral constrained \underline{Co}ntinued \underline{Fr}action \underline{Ge}nerative \underline{Net}works (Spec-CoFrGeNet), a novel reparameterization of continued fraction networks which does not rely on non-smooth $\epsilon$-like clamps. 
We take inspiration from physics where the (0,0) element of a Krylov basis Hamiltonian forms a continued fraction, as shown in \textbf{Fig.\,\ref{fig:approach}} and demonstrate the effectiveness of our approach by replacing the  feedforward networks of standard architectures since, the previous work \citep{dhurandhar2026cofrgenet} showed the best results with such a replacement. We summarize \textbf{our contributions} as follows:

$\bullet$ We introduce a novel parameterization of continued fraction networks, which constrains the spectrum of an associated tridiagonal matrix, thus absolving the need for a non-smooth clamp that can make training and inference unstable even with incremental training. 

$\bullet$ We perform experiments with three different LLM families, namely GPT-2 XL, Granite 4.1 and Llama-MoE.
Across multiple evaluation datasets, \name outperforms the previous continued fraction approach. \name also outperforms standard training of these models in many cases, while using $\frac{2}{3}$ to $\frac{1}{2}$ the parameters. 

$\bullet$ We show that our new approach makes continued fraction networks amenable to even diffusion models for vision, where the previous approach is considerably worse.

\vspace{-4mm}
\section{Related Work}
\vspace{-2mm}
The advent of the Transformer architecture \citep{attention2017} has played a significant contribution in the large scale growth of modern generative models. One defining part of this architecture is the self-attention which models interactions across tokens. Another important part is the FFN, which performs nonlinear transformation on each token. In fact, FFNs occupy a substantial amount of Transformer parameters. Thus existing research trying to improve the efficiency and expressivity of FFNs fall under 3 buckets. The first direction aims to create alternative activation functions and gating mechanisms, such as GELU~\citep{hendrycks2016gelu}, Swish/SiLU~\citep{ramachandran2017searching}, GLU~\citep{shazeer2020glu}, as well as the widely adopted SwiGLU \citep{touvron2023llama,touvron2023llama2,grattafiori2024llama3}.
The second direction aims at changing computation through sparse mixture-of-experts (MoE), which routes inputs selectively to expert networks \citep{shazeer2017outrageously,lepikhin2021gshard,fedus2022switch,jiang2024mixtral,dai2024deepseekmoe,zhu2024llamamoe}. The third direction aims to change the function class itself. These include Kolmogorov--Arnold Networks (KANs)~\citep{liu2024kan}, which consists of learnable univariaye functions, polynomial networks~\citep{chrysos2020pinets} which leverage higher-order polynomial interactions as well as Monarch~\citep{dao2022monarch}. However, scalability, optimization behavior, efficiency, remain import open coniderations. 
Our closest work is CoFrGeNets \citep{dhurandhar2026cofrgenet}, which propose continued fraction based architectures discussed in Section\,\ref{sec: intro} and further discussed in Section\,\ref{sec: background}. 

Diffusion transformers constitute a major class of generative models for image generation. DiT \citep{peebles2023scalable} established the Transformer architecture as a competitive backbone retaining the aforementioned MLP sublayers and attention blocks. Further developments have taken place based on the objective function such as adopting flow-based losses that learn a velocity field \citep{ma2024sit} as well as introducing an auxillary representation aligning objective \citep{yu2024representation}. Finally DDT \citep{wang2026ddt} introduced modern FFN designs such as SwiGLU, while decoupling the network into an encoder-decoder block. 

In this work, we replace the FFNs in both language and vision models with a gated CoFrNet architecture shown in \textbf{Fig. \ref{fig:mlp}}, but where each layer is reparametrized as described in Section \ref{sec: method}.

\section{Background}
\label{sec: background}

\paragraph{Continued fraction generative networks.}
In \citet{dhurandhar2026cofrgenet}, the conventional transformer components are replaced by ensembles of the aforementioned continued fraction ladders (see \textbf{Sec.\,\ref{sec: intro}}). 
A pictorial representation of the architecture is shown in \textbf{Fig. \ref{fig:mlp}} in the appendix.
The inputs to these ladders, i.e., the partial denominators of the continued fractions, are provided by learned affine functions. Multiple such ladders are then combined through a linear projection. The \textbf{depth} of each ladder is the depth of the corresponding continued fraction, while the number of parallel ladders in the ensemble is called the \textbf{width}. Additionally, \citet{dhurandhar2026cofrgenet} adopt an \textbf{incremental training} process, where the parameters of the continued fraction network are optimized in a progressive fashion, i.e., parameters at depth $i$ are optimized only during the final
$t/2^i$ iterations of a run containing $t$ total iterations. 

The FFNs in Transformer architectures have two features that are relevant to their replacement by continued fraction ladders. The first is a \textbf{gating} operation. Given an input $x\in\mathbb{R}^{p}$, two linear projections are computed, one of which passes through a non-linear activation (viz.~Swish, SwiGLU, etc.). Gating refers to the element-wise multiplication of these two projections. Continued fraction replacements of FFNs unconditionally use such gating. 
% \SP{what else, gating?}
The second feature is \textbf{expansion}, referring to how FFNs first project each token from the model dimension ($d_{\mathrm{model}}$) to a higher-dimensional representation ($d_{\mathrm{mlp}}$), and then project the representation back to the model dimension. We define the \textbf{expansion factor} $\alpha$ as the ratio $d_{\mathrm{mlp}} / d_{\mathrm{model}}$. While typical FFN architectures are expanded, i.e., $\alpha > 1$, usually CoFrNets adopt a non-expanded representation, i.e., $\alpha = 1$ to save parameters. 

\paragraph{From Hamiltonians to continued fractions.}
The Hamiltonian operator plays a central role in the study of quantum physics. Not only does it govern the time evolution of a quantum system, but it also characterizes the energy structure of such a system \citep{sakurai1994modern}. 
Indeed, a central challenge in quantum physics is to compute the spectrum of the Hamiltonian, i.e., solve $H|\psi_n\rangle = E_n|\psi_n\rangle$, where $E_n$ are the energy eigenvalues and $|\psi_n\rangle$ are the corresponding energy eigenstates. 
However, for large quantum systems, computing the full spectrum is computationally prohibitive and motivates the use of iterative methods such as the Lanczos algorithm, which can efficiently compute a number of extremal eigenvalues and eigenstates without the full decomposition \citep{lin1990exact,dagotto1994correlated}. 
The Lanczos algorithm creates a Krylov subspace through the repeated application of the Hamiltonian to an initial state \citep{dagotto1994correlated,saad2011numerical}. An $m$-dimensional Krylov subspace is thus defined by 
\[
\mathcal{K}_m(H,|\phi_0\rangle)
=
\operatorname{span}
\left\{
|\phi_0\rangle,
H|\phi_0\rangle,
H^2|\phi_0\rangle,
\ldots,
H^{m-1}|\phi_0\rangle
\right\}.
\]
Subsequently, the Krylov basis vectors are orthonormalized iteratively to construct an orthonormal basis
% Subsequent basis states are created in an iterative fashion. The Krylov basis vectors are then converted into an orthonormal basis
$\{\phi_n\}_{n=0}^{m-1}$ as follows:
\[
\begin{aligned}
&\beta_{n+1}|\phi_{n+1}\rangle
=
H|\phi_n\rangle
-
\alpha_n|\phi_n\rangle
-
\beta_n|\phi_{n-1}\rangle \\
&\alpha_n
=
\langle \phi_n | H | \phi_n \rangle,
\qquad
\beta_n
=
\langle \phi_{n-1} | H | \phi_n \rangle.
\end{aligned}
\]
In this constructed basis, the Hamiltonian is projected into the Krylov subspace and its matrix representation becomes the following symmetric tridiagonal matrix \citep{dagotto1994correlated}: 
\[
T_m =
\begin{bmatrix}
\alpha_0 & \beta_1 & & \\
\beta_1 & \alpha_1 & \ddots & \\
& \ddots & \ddots & \beta_{m-1} \\
& & \beta_{m-1} & \alpha_{m-1}
\end{bmatrix}.
\]

Meanwhile, the resolvent of the Hamiltonian, given by $(z-H)^{-1}$ ($z$ is a complex parameter with the units of energy), is another important quantity frequently used in various domains of quantum mechanics. The matrix elements of the resolvent defines the Green's function. Using the $m$-dimensional Krylov subspace defined above, the (0,0)th element of the matrix representation of the Green's function can be approximated as:
\[
G_{00}(z)
\approx
% \langle \phi_0 | (z-H)^{-1} | \phi_0 \rangle
[(zI-T_m)^{-1}]_{00}
=
\cfrac{1}{
z-\alpha_0
-
\cfrac{\beta_1^2}{
z-\alpha_1-\ddots-
\cfrac{\beta_{m-1}^2}{
z-\alpha_{m-1}
}
}
},
\]
which we recognize as a {continued fraction}. This (0,0) element can be easily computed with recursive elimination with linear complexity.

Additionally, the above inverse also admits another form as follows:
\[
(zI-T_m)^{-1}
= \sum_{n=0}^{m-1} \frac{|\psi_n\rangle\langle \psi_n|}{z-E_n}
\] where $\psi_n$ and $E_n$ are the eigenvectors and eigenvalues of $T_m$.

\section{Methodology}
\label{sec: method}
We use the end results of the progression of ideas presented in Section \ref{sec: background} as the basis of our proposed method. The crux is %of our method lies on 
the following result: the poles of $G_{00}(z)$ are exactly the eigenvalues of $T_m$
\citep{economou2006green}. 
Note that although the development in Section \ref{sec: background} comes from quantum mechanics, the consequences are actually linear algebraic, i.e., they can be independent of the ``quantum nature.'' As a result, for our purposes we can freely set $\beta_i = 1$, $\alpha_i= a_i \in \mathbb{R}$, and $z=0$. Thus our continued fraction reduces from $G_{00}(z)$ to $
f(a)
=
\cfrac{1}{
a_1
-
\cfrac{1}{
a_2 - \cdots
}
}
$, where the corresponding tridiagonal matrix is $
T(a)=
\begin{bmatrix}
a_1 & 1   &        &        \\
1   & a_2 & \ddots &        \\
    & \ddots & \ddots & 1    \\
    &        & 1      & a_d
\end{bmatrix}
 $. %The poles of $f(a)$ are exactly the eigenvalues of the matrix $T(a)$. 

Recall our motivation from Section 1, 
 % TODO: section hyperlink
where we wanted to avoid poles of the continued fractions for stable training, without the reliance on external parameters like $\epsilon$ (see \eqref{eq:eps}). 
Since we have set $z = 0$ above, we wish to avoid poles at $z = 0$. This is equivalent to $T(a)$ not having any zero eigenvalues, i.e., it is a non-singular matrix.

\paragraph{Sufficient condition.}
To derive a sufficient condition for such non-singularity, we leverage the Gershgorin circle theorem \citep{horn2012matrix} which states the following : For a matrix $A\in\mathbb{C}^{d\times d}$, every eigenvalue lies in at least one disk
\[
\mathcal{D}_i(A)
=
\left\{
z\in\mathbb{C}:
|z-A_{ii}|\leq R_i(A)
\right\},
\qquad
R_i(A)
=
\sum_{j\neq i}|A_{ij}|.
\]

\begin{wrapfigure}{r}{.69\textwidth}
\vspace{-.12cm}
% \begin{figure}
\centering
\includegraphics[width=0.69\textwidth]{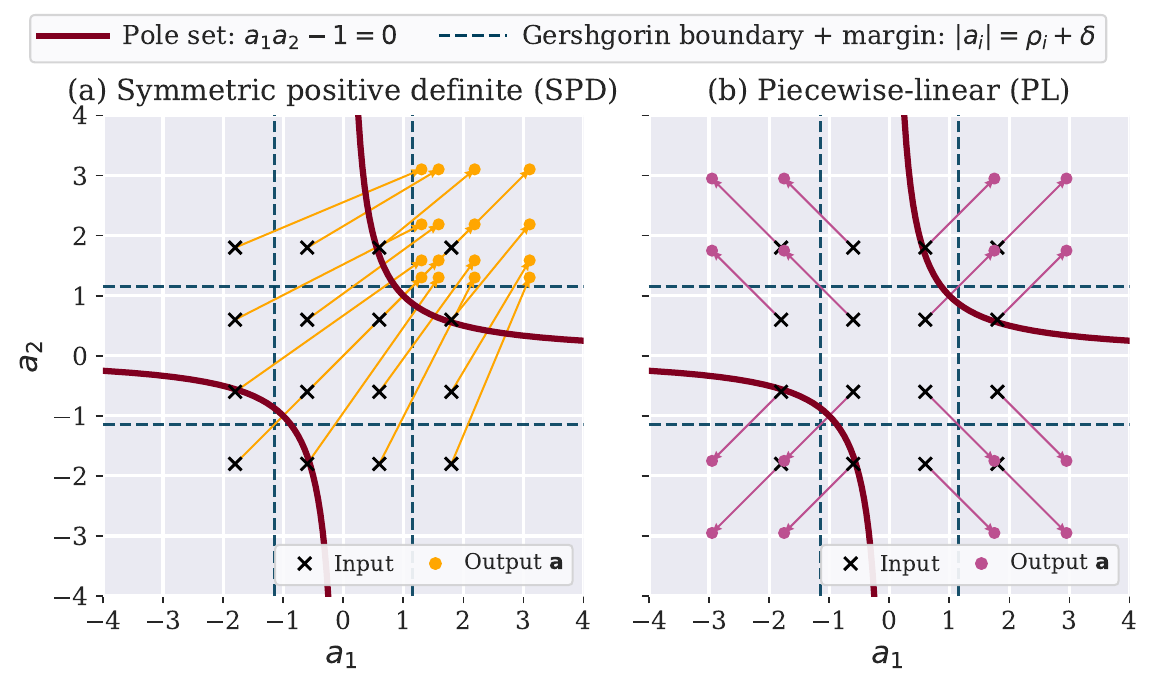}
\vspace{-.7cm}
    \caption{\small{Our proposed parameterizations successfully avoid poles by moving the inputs away from the Gershgorin boundary, as visualized with two-dimensional datapoints. `Input' refers to values of $w_i^Tx$ in \eqref{eq:softplus_param} and \eqref{eq:piecewise_param}. $[a_1,a_2]$ refers to the resulting parameterized values.}}
    \label{fig:parameterization}
% \end{figure}
\vspace{-1.5cm}
\end{wrapfigure}

For the tridiagonal matrix $T(a)$, Gershgorin disks have the radii  $\rho_i$, given by:
\[
\begin{aligned}
\rho_i
&= \sum_{j\neq i}|T_{ij}| \\
&= \begin{cases}
1, & i\in\{1,d\},\\[4pt]
2, & 1<i<d.
\end{cases}
\end{aligned}
\]
and are centered at $a_i$. Thus a sufficient condition for avoiding poles is to simply push each disk away from the origin (i.e. the zero eigenvalue) by its radius: 
\begin{equation}
\begin{aligned}
|a_i|>\rho_i,\\
\qquad i=1,\ldots,d,
\label{eq:condition}
\end{aligned}
\end{equation}

\paragraph{Reparameterizations.}

In \cite{dhurandhar2026cofrgenet,puri2021cofrnets}, the partial denominator scalars are parameterized as the inner product of a learnable weight vector and the input. Following this, we enforce the condition in \eqref{eq:condition} by reparameterizing the partial denominators as
\begin{equation}
\boxed{
a_i(x)
=
\rho_i
+
\operatorname{softplus}(w_i^T x)
+
\delta, \qquad
\delta>0}
\label{eq:softplus_param}
\end{equation}
where softplus is a smooth strictly positive mapping given by $\operatorname{softplus}(z)=\log(1+e^z)$, and $\delta$ is a small margin, set as $\delta=10^{-3}$ in our experiments. This leads to a symmetric positive definite (SPD) $T(a)$ because all of its eigenvalues are positive.

Alternatively, we can preserve the sign of $w_i^T x$ by the following piecewise linear (PL) parameterization: 
\begin{equation}
a_i(x)
=
\begin{cases}
\rho_i+ w_i^T x+ \delta,
& w_i^T x \geq 0,\\[4pt]
-\rho_i+ w_i^T x -\delta,
& w_i^T x<0.
\end{cases}
\label{eq:piecewise_param}
\end{equation}
\emph{Note that the $\delta$ here is a small offset at the continued fraction \textbf{input} to enforce a condition in the spectral domain (i.e. avoid zero eigenvalues), and not a non-smooth clamp like $\epsilon$ that directly determines the scale of the continued fraction \textbf{output}.} The latter is something that has to be carefully tuned, while the former can be set to a fixed value much smaller than $\epsilon$ (i.e. $10^{-3}$) across all our experiments, while still obtaining good results.

We explore the effects of the proposed parameterizations for a depth-$2$ CoFrNet in \textbf{Fig.~\ref{fig:parameterization}}. Let us consider $\mathbf{a}=[a_1,a_2]$ to be the parameterized values. 
At depth 2, the matrix $T(a)$ becomes $\begin{bmatrix}
a_1 & 1\\
1 & a_2
\end{bmatrix}$. The pole set corresponds to values of $(a_1,a_2)$ for which $T(a)$ has a zero eigenvalue, i.e., $\det T(a)=0 \implies a_1a_2=1$. For this $2\times 2$ $T(a)$, the Gershgorin radius is 1 and from \eqref{eq:condition}, the sufficient condition becomes $|a_i|>1$. Thus, the dashed lines in \textbf{Fig.~\ref{fig:parameterization}} denote the margin-adjusted Gershgorin boundaries $|a_i|=1+\delta$ beyond which all points are pushed.
Thus we can see from the illustration how our parameterizations avoid the complete pole set (the hyperbolas). 
However, the sign-dependent shift makes
PL discontinuous whenever either $a_i$ crosses zero. In contrast, SPD maps all inputs continuously through the softplus transformation, although it is restricted to be positive. This smoothness however favors optimization and is therefore our preferred parametrization in the experiments.

\section{Experiments}
\label{sec: experiment}
In this section, we evaluate \name across both language and image generation tasks, allowing us to establish the generality of our approach. We \textbf{replaced the feedforward network} of all the models used, with the CoFrGeNet architecture \citep{dhurandhar2026cofrgenet} along with the modifications of \name (see \textbf{Fig.\,\ref{fig:mlp}} in Appendix\,\ref{app: architecture}).   

\begin{wraptable}{r}{.4\textwidth}
% \begin{table}
\vspace{-0.3cm}
% \scriptsize
\centering
\caption{\small{Below we see architectural details such as number of parameters, number of layers, embedding dimension ($p$) and expansion factor ($\alpha$) of the different (original) models.}}
% \vspace{2mm}
\resizebox{0.4\textwidth}{!}{
\begin{tabular}{c|c|c|c|c}
        \toprule
        \textbf{Model} &
        \textbf{Size} &
        \textbf{Layers} &
        \textbf{$p$} &
        \textbf{$\alpha$} \\
        \midrule

        GPT-2 XL & 1.5B & 48 & 1600 & 4 \\
        Granite 4.1 & 2.14B & 24 & 2560 & 3.2 \\
        Llama-MoE & 403M & 30 & 576 & 2.67 \\

        \bottomrule
    \end{tabular}
    }
 \label{tab:archdet}
 \vspace{-0.3cm}
\end{wraptable}
% \end{table}

 % \vspace{-4mm}
\definecolor{lightgray}{RGB}{242,242,242}
\definecolor{lightblue}{RGB}{232,241,248}
\begin{table}[b]
    \centering
    \caption{\small{Perplexity of pretrained \textbf{GPT-2 XL} models on various evaluation datasets (lower is better). All CoFrGeNets have the \textbf{FNG} setting. The standard models do not have gating.  $\dagger$ denotes the standard model trained with an expanion factor of 1. \protect\colorbox{lightgray}{\strut Light gray} denotes the default standard setting, while \protect\colorbox{lightblue}{\strut light blue} denotes our proposed method. The best results are highlighted in \textbf{bold}.}}
    \label{tab: gpt}
    \setlength{\tabcolsep}{7pt}
    \renewcommand{\arraystretch}{1.15}
    \resizebox{\linewidth}{!}{%
    \begin{tabular}{c|c|c|c|c|c|c|c|c}
        \toprule
        \toprule
        \textbf{Dataset} &
        \textbf{Method} & 
        % \textbf{\shortstack{Expansion\\ratio}} 
        % \textbf{$\alpha=1$} 
        % &
        \textbf{Params} &
        \textbf{PTB} &
        \textbf{Wikitext2} &
        \textbf{Lambada} &
        \textbf{AgNews} &
        \textbf{LM1B} &
        \textbf{Wikitext103} \\
        \midrule

        \multirow{4}{*}{OWT}
        & \cellcolor{lightgray} Standard
        & \cellcolor{lightgray} 1.5B
        & \cellcolor{lightgray} 30.12
        & \cellcolor{lightgray} 18.30
        & \cellcolor{lightgray} 8.66
        & \cellcolor{lightgray} 37.13
        & \cellcolor{lightgray} 41.20
        & \cellcolor{lightgray} 17.50 \\
        \cline{3-9}
        \noalign{\vskip 1.5pt}

        & Standard $^{\dagger}$ 
        & 
        & 35.98 
        & 27.74
        & 16.35
        & 50.75
        & 51.95 
        & 29.29\\
        & CoFrGeNet 
        &  985M  
        & 29.89 
        &\bf 17.12
        & 8.12
        & 35.72
        & \bf 40.14 
        & 16.14\\

        & \cellcolor{lightblue} \name
        & \cellcolor{lightblue} 
        & \cellcolor{lightblue} \textbf{29.87}
        & \cellcolor{lightblue} 17.14
        & \cellcolor{lightblue} \bf 8.11
        & \cellcolor{lightblue} \bf 35.69
        & \cellcolor{lightblue} 40.17
        & \cellcolor{lightblue} 16.13 \\

        \midrule

        \multirow{4}{*}{GW}
        & \cellcolor{lightgray} Standard
        & \cellcolor{lightgray} 1.5B
        & \cellcolor{lightgray} \bf 29.07
        & \cellcolor{lightgray} 19.12
        & \cellcolor{lightgray} 31.78
        & \cellcolor{lightgray} 45.62
        & \cellcolor{lightgray} 52.36
        & \cellcolor{lightgray} 18.93 \\
        \cline{3-9}
        \noalign{\vskip 1.5pt}

        & Standard $^{\dagger}$ 
        & 
        & 34.26
        & 27.38
        & 39.44
        & 61.81
        & 58.64
        & 29.02
        \\

        & CoFrGeNet 
        &  985M  
        & 29.72
        & 18.13
        & \bf 30.52
        & 41.63
        & \bf 46.83
        & 18.11
        \\

        & \cellcolor{lightblue} \name
        & \cellcolor{lightblue}
        & \cellcolor{lightblue} 29.68
        & \cellcolor{lightblue} \bf 18.10
        & \cellcolor{lightblue} 30.55
        & \cellcolor{lightblue} \bf 41.61
        & \cellcolor{lightblue} 46.86
        & \cellcolor{lightblue} \bf 18.06
        \\

        \bottomrule
    \end{tabular}
    }
\end{table}

For all experiments, we compare \name with incremental training of CoFrGeNets (henceforth referred to simply as `CoFrGeNet') and also with the original feedforard network based model (henceforth referred to simply as `Standard'). Typically, all CoFrNet variants have an expansion factor of 1, unless otherwise mentioned. Note that the continued fraction based architecture and the expansion factor significantly reduces the number of parameters for CoFrNet variants. Hence for fair comparison, we also consider the original models with an expansion factor of 1 (often marked with a $\dagger$).  For all experiments, $\epsilon=0.1$ in CoFrGeNets, while width and depth are chosen as 5. Models are referred to as FNG (Feedforward Non-expanded Gated) when $\alpha=1$ with gating, or FG (Feedforward Gated), when $\alpha>1$. Results with width and depth 1 are in the appendix. 

\rev{Also note that we tried two other baselines where we forego $\epsilon$ and $\rho$: i) apply softplus to $wx$ in each layer without adding $\rho$ and ii) apply softplus to the entire continuant denominator. $\delta=0.001$ was added to both these baselines in the denominators to prevent divide-by-zero. For language both led to unstable training where the validation loss reached NaNs, as the values got extremely close to $\delta$ in the denominators for certain inputs. For vision too we saw a similar trend.}

\subsection{Language Modeling}

For language modeling, we test \name on three diverse models \gpt, \granite and \llama. Unless otherwise mentioned, all experiments were performed with a per-GPU batch size of 16 using the AdamW optimizer ($\beta_1=0.9, \beta_2=0.95$) and a weight decay of 0.1.

% \subsubsection{Pretraining}

\paragraph{GPT-2 XL.} 
We pre-train GPT-2 XL and the aforementioned baselines on OpenWebText (OWT)~\citep{owt} and GneissWeb-35B (GW)~\citep{gneissweb}. 
For each pretrained model, we report the perplexity on Penn Tree Bank (PTB) \citep{Marcus93_PTB}, Wikitext2 \citep{Merity2017_Wikitext}, Wikitext103 \citep{Merity2017_Wikitext}, Lambada \citep{Paperno2016_LAMBADA}, AgNews \citep{AgNews2015} and One Billion Words (LM1B) \citep{Chelba2014_lm1b}. We use a stride of 512 for WikiText-2
and WikiText-103, following prior evaluation protocols, and a stride of
256 for all other datasets. We use the recommended settings for pretraining \gpt in \url{https://github.com/karpathy/nanoGPT} where, the learning rate is $6\times 10^{-4}$, weight decay is $0.1$, there is no dropout and the maximum iterations is $600$K ($2$K warmup). All models were trained on $16$ H100 GPUs leveraging distributed data parallel (DDP) framework.

In most datasets in \textbf{Table\,\ref{tab: gpt}},  \name outperforms previous CoFrGeNet versions and also performs better than the standard version of the model with a much higher number of parameters. 

% \vspace{-5mm}
\definecolor{lightgray}{RGB}{242,242,242}
\definecolor{lightblue}{RGB}{232,241,248}
\begin{table}[h]
    \centering
    \caption{\small{Downstream accuracies of finetuned \textbf{GPT-2 XL} models on GLUE. All CoFrGeNets have the FNG setting. The standard models do not have gating. $\dagger$ denotes the standard model trained with no expansion.}}
    \label{tab:finetune}
    \setlength{\tabcolsep}{7pt}
    \renewcommand{\arraystretch}{1.15}
    \resizebox{\linewidth}{!}{%
    \begin{tabular}{c|c|c|c|c|c|c|c|c|c|c}
        \toprule
        \toprule
        \textbf{Dataset} &
        \textbf{Model} & 
        \textbf{Params} &
        \textbf{MNLI} &
        \textbf{QQP} &
        \textbf{QNLI} &
        \textbf{SST2} &
        \textbf{COLA} &
        \textbf{MRPC} &
        \textbf{RTE} &
        \textbf{WNLI} 
        \\
        \midrule

        \multirow{4}{*}{OWT}
        & \cellcolor{lightgray} Standard
        & \cellcolor{lightgray} 1.5B
        & \cellcolor{lightgray}86.89 &
        \cellcolor{lightgray}88.93 &
        \cellcolor{lightgray}91.35 &
        \cellcolor{lightgray}93.56 &
        \cellcolor{lightgray}81.78 &
        \cellcolor{lightgray}79.83 &
        \cellcolor{lightgray}60.27 &
        \cellcolor{lightgray}58.28 \\
        \cline{3-11}
        \noalign{\vskip 1.5pt}

        & Standard $^{\dagger}$ &
        & 81.17 & 81.35 & 83.94 & 87.22 & 75.99 & 74.37 & 56.42 & 58.28\\
        & CoFrGeNet 
        &  985M  
        & \bf 87.24 & \bf 89.95 & \bf 91.87 & \bf 94.12 & \bf 82.57 & \bf 80.17 & \bf 61.36 & \bf 58.30\\

        & \cellcolor{lightblue} \name
        & \cellcolor{lightblue} 
        & \cellcolor{lightblue}\bf 87.26 &
            \cellcolor{lightblue}\bf 89.95 &
            \cellcolor{lightblue}\bf 91.89 &
            \cellcolor{lightblue}\bf 94.16 &
            \cellcolor{lightblue}\bf 82.59 &
            \cellcolor{lightblue}\bf 80.21 &
            \cellcolor{lightblue}\bf 61.35 &
            \cellcolor{lightblue}\bf 58.30 \\

        \midrule

        \multirow{4}{*}{GW}
        & \cellcolor{lightgray} Standard
        & \cellcolor{lightgray} 1.5B
        & \cellcolor{lightgray}78.28 &
\cellcolor{lightgray}86.83 &
\cellcolor{lightgray}\bf 82.93 &
\cellcolor{lightgray}91.82 &
\cellcolor{lightgray}74.18 &
\cellcolor{lightgray}77.72 &
\cellcolor{lightgray}60.19 &
\cellcolor{lightgray}\bf 58.33\\
        \cline{3-11}
        \noalign{\vskip 1.5pt}

        & Standard $^{\dagger}$ 
        & 
        & 74.83 & 84.74 & 77.82 & 83.94 & 68.93 & 71.83 & 57.85 & \bf 58.33
        \\

        & CoFrGeNet 
        &  985M  
        & \bf 79.23 & \bf 87.24 & 82.69 & \bf 92.38 & \bf 74.79 & \bf 78.04 & \bf 61.37 & \bf 58.33
        \\

        & \cellcolor{lightblue} \name
        & \cellcolor{lightblue}
        & \cellcolor{lightblue}\bf 79.62 &
\cellcolor{lightblue}\bf 87.26 &
\cellcolor{lightblue}82.73 &
\cellcolor{lightblue}\bf 92.36 &
\cellcolor{lightblue}\bf 74.83 &
\cellcolor{lightblue}\bf 78.01 &
\cellcolor{lightblue}\bf 61.35 &
\cellcolor{lightblue}\bf 58.33
        \\

        \bottomrule
    \end{tabular}
    }
\end{table}
We also evaluate finetuning performance on the pretrained \gpt models by finetuning on GLUE \citep{glue} (classification) tasks and compare accuracies as done in previous works \citep{dhurandhar2026cofrgenet}. For finetuning, we choose the best performing learning rates for all the baselines, thus leading to $0.125\times 10^{-4}$ for \name and $0.25\times 10^{-4}$ for all other methods. Finetuning was achieved with a single A100 GPU for each model. As observed in \textbf{Table\,\ref{tab:finetune}}, \name achieves better performance in most downstream tasks, when compared to Standard.

\paragraph{Granite 4.1.}
We pretrain Granite 4.1 on 125B tokens drawn from web and mathematical datasets in NVIDIA's Nemotron data repository, including Web-Nemotron-CC, MegaMath-Web, and FineMath-3Plus. 
We evaluate these pretrained models on OpenBookQA~\citep{OpenBookQA2018}, PIQA~\citep{piqa},
ARC-Easy~\citep{arc}, WinoGrande~\citep{winogrande},
HellaSwag~\citep{hellaswag}, LAMBADA (OpenAI)
~\citep{radford2018improving}, BoolQ~\citep{boolq}, and
SciQ~\citep{SciQ}. These benchmarks cover open-domain and
commonsense question answering, physical and commonsense reasoning,
and language understanding.
% \vspace{-2mm}

We pre-train Granite 4.1 for $30$K iterations, with $3$K warmup steps. The training was performed using the Adam optimizer with an
initial learning rate of $3\times 10^{-4}$ and an annealing schedule with no dropout.
All Granite 4.1 models were trained using 64 H100 GPUs with FSDP.

\definecolor{lightgray}{RGB}{242,242,242}
\definecolor{lightblue}{RGB}{232,241,248}
\begin{table}[h]
    \centering
    \caption{\small{Zero-shot accuracies on open domain Q\&A, reasoning and text understanding tasks for \textbf{Granite 4.1} and its baselines.}}
    \vspace{0.4pt}
    \label{tab: granite}
    \setlength{\tabcolsep}{7pt}
    \renewcommand{\arraystretch}{1.15}
    \resizebox{\linewidth}{!}{%
    \begin{tabular}{c|c|c|c|c|c|c|c|c|c|c}
        \toprule
        \toprule
        \textbf{Method} & \textbf{Setting} &

        \textbf{Params} &
        \textbf{OpenBookQA} &
        \textbf{PIQA} &
        \textbf{ARC-Easy} &
        \textbf{WinoGrande} &
        \textbf{HellaSwag} &
        \textbf{LAMBADA} &
        \textbf{BoolQ} &
        \textbf{SciQ}\\
        \midrule

        Standard & \multirow{3}{*}{FG} & \multirow{3}{*}{2.14B}  & 35.2 & \bf 74.8 & 69.3 & 56.9 & \bf 59.4 & 48.7 & 57.7 & \bf 90.8 \\
        CoFrGeNet & & & 37.2 & 74.4 & 69.9 & 58.1 & 59.1 & 48.6 & 60.9 & 90.2 \\
        Spec-CoFrGeNet &   & & \cellcolor{lightblue}\textbf{37.4}
& \cellcolor{lightblue}74.6
& \cellcolor{lightblue}\textbf{70.0}
& \cellcolor{lightblue}\textbf{58.4}
& \cellcolor{lightblue}\textbf{59.4}
& \cellcolor{lightblue}\textbf{49.0}
& \cellcolor{lightblue}\textbf{61.2}
& \cellcolor{lightblue}90.0
 \\
    
        \cline{2-11}
        \noalign{\vskip 1.5pt}

        Standard & \multirow{3}{*}{FNG} & \multirow{3}{*}{1.1B} & 34.9 & 72.9 & 63.9 & 55.7 & 52.7 & 44.7 & 57.2 & 86.1 \\
        CoFrGeNet & & & 35.2 & 73.1 & 63.9 & 55.9 & 52.7 & 45.0 & 59.1 & 86.5\\
        Spec-CoFrGeNet &  & & \cellcolor{lightblue}35.4
& \cellcolor{lightblue}73.1
& \cellcolor{lightblue}64.1
& \cellcolor{lightblue}55.8
& \cellcolor{lightblue}53.2
& \cellcolor{lightblue}45.0
& \cellcolor{lightblue}60.2
& \cellcolor{lightblue}86.7
\\

        \bottomrule
    \end{tabular}
    }
\end{table}
\definecolor{lightgray}{RGB}{242,242,242}
\definecolor{lightblue}{RGB}{232,241,248}
\begin{table}[t]
    \centering
    \caption{\small{Zero-shot accuracies on open domain Q\&A, reasoning and text understanding tasks for \textbf{Llama-MoE} and its baselines. All CoFrGeNets have the \textbf{FNG} setting. $\dagger$ denotes the standard model trained with no expansion (i.e. $\alpha=1$).}}
    \vspace{0.2pt}
    \label{tab: llama}
    \setlength{\tabcolsep}{7pt}
    \renewcommand{\arraystretch}{1.15}
    \resizebox{\linewidth}{!}{%
    \begin{tabular}{c|c|c|c|c|c|c|c|c|c}
        \toprule
        \toprule
        \textbf{Model} & 
        % \textbf{\shortstack{Expansion\\ratio}} 
        % \textbf{$\alpha=1$} 
        % &
        \textbf{Params} &
        \textbf{OpenBookQA} &
        \textbf{PIQA} &
        \textbf{ARC-Easy} &
        \textbf{WinoGrande} &
        \textbf{HellaSwag} &
        \textbf{LAMBADA} &
        \textbf{BoolQ} &
        \textbf{SciQ}\\
        \midrule

        \cellcolor{lightgray} Standard
        & \cellcolor{lightgray} 403M
        & \cellcolor{lightgray} 14.2
        & \cellcolor{lightgray} 60.0
        & \cellcolor{lightgray} 47.2
        & \cellcolor{lightgray} \bf 50.1
        & \cellcolor{lightgray} 27.5
        & \cellcolor{lightgray} 17.8
        & \cellcolor{lightgray} \bf 60.8
        & \cellcolor{lightgray} 72.6 \\
        \cline{2-10}
        \noalign{\vskip 1.5pt}

        Standard $^{\dagger}$ 
        & 
        & 14.2
        & 57.6
        & 41.7
        & 49.9
        & 26.7
        & 15.7
        & 60.7
        & 68.1 \\
        
        CoFrGeNet 
        & 204M
        & 15.2
        & 61.0
        & 51.1
        & 49.8
        & \bf 28.1
        & \bf 21.2
        & 53.2
        & 77.6 \\

        \cellcolor{lightblue} \name
        & \cellcolor{lightblue} 
        & \cellcolor{lightblue} \bf 15.4
        & \cellcolor{lightblue} \bf 61.2
        & \cellcolor{lightblue} \bf 51.2 
        & \cellcolor{lightblue} \bf 50.1 
        & \cellcolor{lightblue} 28.0 
        & \cellcolor{lightblue}  21.1
        & \cellcolor{lightblue}  53.7
        & \cellcolor{lightblue}  \bf 78.0
        \\

        \bottomrule
    \end{tabular}
    }
\end{table}

As observed in \textbf{Table \,\ref{tab: granite}}, the non-expanded versions contain almost half the parameters of the expanded versions yielding competitive but worse performance. Hence, we consider the expanded versions of CoFrGeNet and \name for Granite 4.1. \name outperforms the other baselines in six out of eight datasets, thus establishing the success of our approach.

\paragraph{Llama-MoE.}

We pre-train Llama-MoE (configured with 4 experts and one active expert per token) using the 2T token data mixture, which were used to train the Docling \citep{Docling}  models. This mixture contains web data (DCLM2 and DCLM3Plus~\citep{li2024datacomplm}), multilingual data (FineWeb-2-Edu~\citep{fineweb-edu}), code (StarCoder and Stack-Edu~\citep{stack-edu-finemath}), mathematical data (FineMath~\citep{stack-edu-finemath}, InfiWebMath~\citep{infimath}, OPC-FineWeb-Math-Corpus~\citep{opc}), and synthetic data (Cosmopedia~\citep{cosmopedia}). The exact proportions of the different datasets are provided in the Appendix\,\ref{app: setup}. The evaluation setup remains the same as Granite 4.1. 
This was pretrained for about 2M iterations with $2$K warmup steps. All other experimental details remain consistent with Granite 4.1.

As observed in \textbf{Table\,\ref{tab: llama}}, even though the CoFrGeNet models have almost half the number of parameters when compared to the standard model, \name outperforms all baselines in five out of eight tasks. 
In the other tasks, it maintains competitive performance with the CoFrGeNet model trained with incremental training.

% \begin{table}[t]
\begin{wraptable}{r}{.7\textwidth}
\vspace{-.4cm}
    \centering
    \caption{\small{Efficiency metrics namely throughput (given in tokens/ day), training time (days) and number of floating point operations (GFLOPs) per token for all models. For GPT2 XL, training time is presented as (OWT / GW)}. $\dagger$ denotes the standard model trained with no expansion (i.e. $\alpha=1$), while $\S$ denotes \name trained with expansion (FG).}
    \label{tab:llm_efficiency}
    \setlength{\tabcolsep}{7pt}
    \renewcommand{\arraystretch}{1.15}
    \resizebox{0.7\textwidth}{!}{%
    \begin{tabular}{c|c|c|c|c|c}
    \toprule
    \textbf{Model} &
    \textbf{Method} &
    \shortstack{\textbf{Parallelism}\\\textbf{Strategy}} &
    %\shortstack{\textbf{Training}\\
    \textbf{Throughput}  &
    \shortstack{\textbf{Training}\\\textbf{Time}} &
    \textbf{FLOPs}\\
    \midrule

     &
    \cellcolor{lightgray} Standard &
       &
    \cellcolor{lightgray} 1.15B &
    \cellcolor{lightgray} 7.9 / 17.2 &
    \cellcolor{lightgray}  9.1 \\
    
     GPT-2 XL &
     Standard$^\dagger$ &
     DDP &
    1.21B &
     7.6 / 16.1 &
     6.1 \\

     &
    \cellcolor{lightblue} \name &
      &
    \cellcolor{lightblue} 1.20B &
    \cellcolor{lightblue} 7.6 / 16.2 &
    \cellcolor{lightblue} 6.1 \\
    
    \midrule

  \multirow{4}{*}{Granite 4.1}  & 
    \cellcolor{lightgray} Standard &
      \multirow{4}{*}{FSDP} &
    \cellcolor{lightgray} 162B &
    \cellcolor{lightgray} 0.77 &
    \cellcolor{lightgray}  12.7 \\
    
    &
     Standard$^\dagger$ &
      &
    203B &
     0.61 &
     6.8 \\

     &
    \cellcolor{lightblue} \name &
      &
    \cellcolor{lightblue} 202B &
    \cellcolor{lightblue} 0.62 &
    \cellcolor{lightblue} 6.8 \\

    &
    \cellcolor{lightblue} \name $^{\S}$ &
      &
    \cellcolor{lightblue} 160B &
    \cellcolor{lightblue} 0.78 &
    \cellcolor{lightblue} 12.8 \\

    \midrule

     &
    \cellcolor{lightgray} Standard &
       &
    \cellcolor{lightgray} 299B &
    \cellcolor{lightgray} 6.7 &
    \cellcolor{lightgray}  1.1 \\
    
     Llama-MoE &
     Standard$^\dagger$ &
     FSDP &
    310B &
     6.3 &
     0.70 \\

     &
    \cellcolor{lightblue} \name &
      &
    \cellcolor{lightblue} 309B &
    \cellcolor{lightblue} 6.4 &
    \cellcolor{lightblue} 0.71 \\

    \bottomrule
\end{tabular}
}
% \end{table}
 % \vspace{-0.4cm}
 \vspace{-1.0cm}
\end{wraptable}

Finally, we present various efficiency metrics in  \textbf{Table\,\ref{tab:llm_efficiency}}. As can be seen \name is more efficient than the corresponding standard models.

\subsection{Image Generation}
To demonstrate the effectiveness of our approach in vision, we perform class-conditional image generation experiments with Decoupled Diffusion Transformers (DDT) \citep{wang2026ddt} on ImageNet-1K at $256\times256$ resolution. Following \cite{wang2026ddt}, images are center cropped using the ADM preprocessing protocol \citep{dhariwal2021diffusion} and encoded into latent representations using the frozen Stability AI VAE \citep{rombach2022high}. We perform experiments using the DDT-XL/2 model as the primary backbone architecture, which contains 28 transformer blocks, with 22 blocks acting as the encoder and 6 blocks acting as the velocity decoder. %Additionally, we consider the DDT-L/2 model which has 24 transformer blocks, where 20 blocks act as the encoder and 4 as the decoder. These models have a hidden dimension of 1152 and 1024 respectively. 
Representation alignment is performed using REPA \citep{yu2024representation}, which aligns the hidden representation of the output of the eighth encoder block with a frozen DINOv2-B/14 teacher \citep{oquab2024dinov}. 
All other architecture details follow the DDT implementation \citep{wang2026ddt}.

\begin{table}[h]
    \centering
    \caption{Performance of various metrics using DDT-XL/2 at 400000 steps. $\uparrow$ refers to higher the better and $\downarrow$ refers to lower the better.}
    \label{tab:image_generation}
    \setlength{\tabcolsep}{7pt}
    \renewcommand{\arraystretch}{1.15}
    \resizebox{\textwidth}{!}{%
    \begin{tabular}{c|c|c|c|c|c|c}
        \toprule
        \textbf{Method} &
        \textbf{Params} &
        \textbf{FID $\downarrow$} &
        \textbf{sFID $\downarrow$} &
        \textbf{Inception Score $\uparrow$} &
        \textbf{Precision $\uparrow$} &
        \textbf{Recall $\uparrow$} \\
        \midrule

        \cellcolor{lightgray}Standard &
        \cellcolor{lightgray} 678M &
        \cellcolor{lightgray} 1.447 &
        \cellcolor{lightgray} 4.345 &
        \cellcolor{lightgray} 312.594 &
        \cellcolor{lightgray} 0.631 &
        \cellcolor{lightgray}0.801   \\

        Standard $^\dagger$ & 493M & 1.626 & 4.417 & 287.476 & 0.620 & 0.800 \\

        \cellcolor{lightblue} \name &
        \cellcolor{lightblue} 493M &
        \cellcolor{lightblue} 1.606 &
        \cellcolor{lightblue} 4.406 &
        \cellcolor{lightblue} 290.835 &
        \cellcolor{lightblue} 0.621 &
        \cellcolor{lightblue} 0.799 \\

        \bottomrule
    \end{tabular}
    }
\end{table}
% \vspace{-2mm}

 \begin{wrapfigure}{r}{.4\textwidth}
\vspace{-.3cm}
% \begin{figure}
    \centering
    \includegraphics[width=0.3\textwidth]{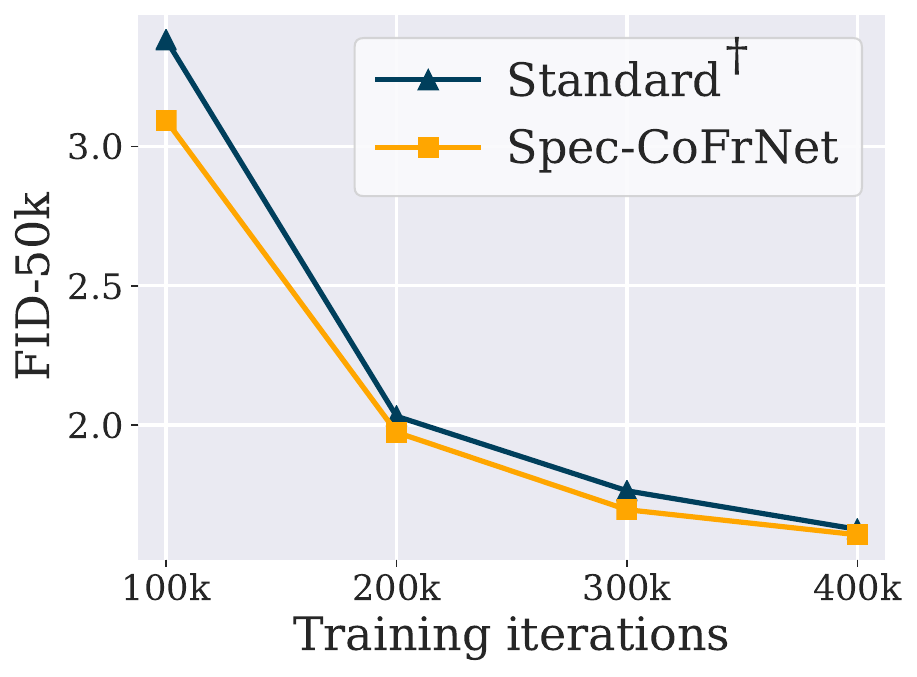}
    \vspace{-.2cm}
    \caption{\small{Convergence of FID measured for 50000 samples generated using DDT-XL/2 models.}}
    \label{fig: fid}
% \end{figure}
% \vspace{-1.4cm}
\end{wrapfigure}

 % \begin{wrapfigure}{r}{.4\textwidth}
% \vspace{-.7cm}
\begin{figure}
    \centering
    \includegraphics[width=\textwidth]{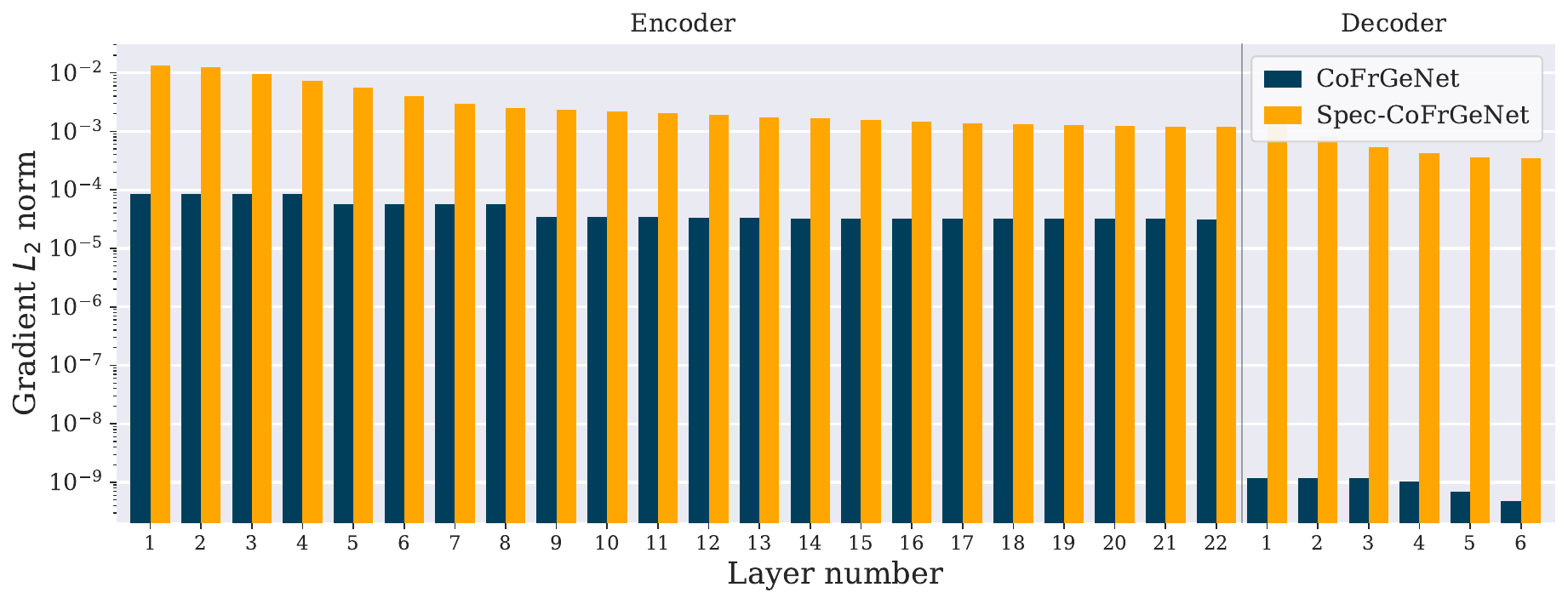}
    \vspace{-.5cm}
    \caption{\small{\rev{Gradient norm comparison for encoder and decoder layers using DDT-XL/2 at 50000 steps just before CoFrGeNet loss becomes NaN. Gradient is computed for loss with respect to output of each layer. As can be seen the gradient norms for CoFrGeNet jump by orders of magnitude from decoder to encoder, a signature of unstable training.}}}
    \label{fig: gradient}
    \vspace{-.1cm}
\end{figure}
% \vspace{-1.4cm}
% \end{wrapfigure}

We train the model using a velocity-based linear flow matching objective \citep{ma2024sit} regularized with a REPA objective. 
The REPA loss is computed using a token-wise negative cosine similarity between projected DDT representations and the corresponding teacher DINOv2-B/14 representations. For classifier-free guidance, class labels are dropped with a probability of 0.1.
An exponential moving average (EMA) of the model, with a decay of 0.9999 was maintained.

% \begin{table}[t]
\begin{wraptable}{r}{.4\textwidth}
\vspace{-1.7cm}
    \centering
    \caption{\small{Efficiency metrics namely throughput (given in images per second per GPU) and training time (in hours) for all variants of DDT-XL/2 models.}}
    \vspace{1mm}
    \label{tab:im_efficiency}
    \setlength{\tabcolsep}{7pt}
    \renewcommand{\arraystretch}{1.15}
    \resizebox{0.4\textwidth}{!}{%
    \begin{tabular}{c|c|c}
    \toprule
    \textbf{Method} &
    \textbf{Throughput} &
    \shortstack{\textbf{Training}\\\textbf{Time}} \\
    \midrule

    \cellcolor{lightgray} Standard &
    \cellcolor{lightgray} 64.78 &
    \cellcolor{lightgray} 54.88 \\

    Standard$^\dagger$ &
    74.79 &
    47.54 \\

    \cellcolor{lightblue} \name &
    \cellcolor{lightblue} 74.57 &
    \cellcolor{lightblue} 48.41 \\

    \bottomrule
    \end{tabular}
    }
% \end{table}
 \vspace{-0.5cm}
\end{wraptable}

These models were trained for 400000 steps following previous works, using AdamW with a constant learning rate of $1 \times 10^{-4}$, $\beta_1=0.9, \beta_2=0.95$ and zero weight decay. %Training was performed with 
We used a per-GPU batch size of 32 using BF16 mixed precision on eight NVIDIA A100s.
%GPUs. 

We evaluate the EMA model by sampling 50,000 class-balanced images, using 250 deterministic Euler steps with the encoder representation recomputed at every step. During sampling, classifier-free guidance is applied for $0<t<0.7$ with a guidance scale of 2.0. Following the ADM evaluation protocol \citep{dhariwal2021diffusion}, we report the FID, sFID, Inception Score, Precision and Recall. FID measures the similarity between real and generated images in the Inception feature space, but it is sensitive to image spatial changes like resizing and preprocessing \cite{leng2026diffusionbench}. sFID provides a complementary metric by capturing spatial relationship. Inception score and Precision characterize sample fidelity while Recall is a proxy for sample diversity.

In our experiments, we observe that the \textbf{CoFrGeNet (with incremental training) cannot learn the input distribution}. Consequently, FID stays stuck at around 565.3 and eventually the loss becomes unstable. This is in fact a much more severe issue of instability, than the one we demonstrated in Section\,\ref{sec: intro} and Fig.\,\ref{fig:epsissue}. \rev{We investigate this issue further through a gradient analysis, where we compute the gradient norms for models trained using both CoFrGeNet and \name. We consider the checkpoints just before the CoFrGeNet training becomes unstable, i.e. the loss becomes NaN. Specifically, we compute $||\partial\mathcal{L}/\partial x_l||_2$, where $\mathcal{L}$ denotes the training loss and $x_l$ denotes the activation after layer $l$ i.e. after residual addition. The loss is computed as the mean loss from 100 randomly selected samples from the dataset. As observed in \textbf{Fig.\,\ref{fig: gradient}}, there is a gradient jump in CoFrGeNet as evidenced by the increase of the gradient norm across the layers especially from the decoder to the encoder where it increased from about $10^{-9}$ to $10^{-4}$, i.e. an increase of the order of $10^{5}$. This happens because the denominators become small and start hitting the $\epsilon$ clamp. In contrast, our method \name shows stable gradient norms indicative of stable training.}
As observed in \textbf{Fig.\,\ref{fig: fid}}, \name converges faster than  non-expanded Standard model. Additionally, from \textbf{Table\,\ref{tab:image_generation}}, we observe that \name performs close to the Standard model while having about $30\%$ less parameters. We also see that our method outperforms the non-expanded standard model in FID, sFID and IS.

\section{Conclusion}
\label{sec:concl}
In summary, inspired by the Green's function and its relation to continued fractions in physics we have provided a novel approach to stably and accurately train CoFrNets that does away with the non-smooth and rather artificial clamping. The trick is to reparametrize CoFrNets so as to spectrally constrain them that leads to a non-singular tridiagonal matrix inverse formed from the CoFrNet layers. Empirically, 
%The benefits of this were seen experimentally, where 
we obtain superior results to CoFrGeNets on language in many cases and significantly better results on vision foregoing the need for incremental training.

Our general idea of simply requiring a non-singular tridiagonal matrix formed from the CoFrNet layers opens up avenues for other reparametrizations. For instance, one could design other functions for each entry or even reparametrize a combination of entries as those would be weaker conditions for invertibility. Additionally, even though we consider generative modeling, our proposed reparameterization is general and can be applied in any discriminative learning task. Such explorations would be interesting to conduct in the future.

\subsection*{AI use statement}

In this work, we used generative AI tools for assistance in coding and assistance in preparing the manuscript such as polishing the writing. We take responsibility for the final content of this work, including text, claims or artifacts produced with the aid of generative AI.

\subsection*{Ethics statement}

Our work provides a novel parameterization for training and we are not releasing a new model or dataset. Thus the risks associated with the method are similar to training any generative modeling architecture and depend on the dataset as well as downstream deployment. All existing safety measures and guardrails applied to other models, are applicable to our method as well.   

\subsection*{Reproducibility statement}

We have provided the details of the training procedure as well as architectural details in the manuscript to facilitate reproducibility. Code is available at \url{https://github.com/IBM/CoFrNets/}.

\bibliography{iclr2027_conference}
\bibliographystyle{iclr2027_conference}

\appendix
\appendix

\counterwithin{figure}{section}
\counterwithin{table}{section}
\counterwithin{equation}{section}

\renewcommand{\theHfigure}{appendix.\thesection.\arabic{figure}}
\renewcommand{\theHtable}{appendix.\thesection.\arabic{table}}
\renewcommand{\theHequation}{appendix.\thesection.\arabic{equation}}

\section{Architecture details}
\label{app: architecture}

\begin{figure}[htbp]
%\vspace{-1cm}
\centering
\includegraphics[width=.5\textwidth]{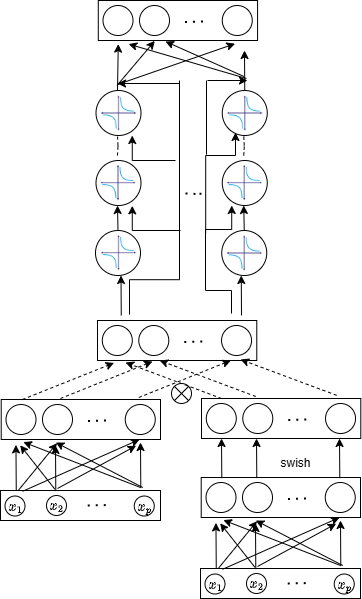}
\caption{CoFrGeNet-FNG architecture replacing FFNs in a transformer block. We create a gated \emph{non-expanded} (i.e. $\alpha=1$) representation that we pass to the CoFrNet ladders. For each of the continued fraction ladders, we use our parameterization from \eqref{eq:softplus_param}, thus forming \name.}
 \label{fig:mlp}
%\vspace{-.8cm}
\end{figure}

\section{Additional dataset details}
\label{app: setup}

\paragraph{Dataset for pretraining Granite 4.1.}

We construct the data mixture from four datasets \texttt{web-nemotron-cc-hq-p2\_0}, \texttt{web-nemotron-cc-hq-p2\_1}, \texttt{megamath-web-pro\_0}, \texttt{finemath-3plus-rewritten\_0} with sampling ratios of 0.35, 0.30, 0.20 and 0.15 respectively.

% Datasets = web-nemotron-cc-hq-p2\_0, web-nemotron-cc-hq-p2\_1, megamath-web-pro\_0, finemath-3plus-rewritten\_0

% Weights = 0.35, 0.30, 0.20, 0.15
\paragraph{Dataset for pretraining Llama-MoE.}

The Docling data mixture contains \texttt{dclm3plus}, \texttt{dclm2}, \texttt{cosmopedia\_v2}, \texttt{fineweb\_math\_corpus}, \texttt{finemath-4plus}, \texttt{infiwebmath-4plus}, \texttt{finemath-3plus}, \texttt{infiwebmath-3plus}, \texttt{starcoder}, \texttt{stack-edu}, \texttt{fineweb-2-edu} with corresponding sampling weights of 0.5, 0.24, 0.06, 0.01, 0.015, 0.01, 0.015, 0.005, 0.01, 0.17 and 0.01 respectively.

\rev{\section{Spec-CoFrGeNet Generations}}
\rev{\paragraph{Granite 4.1}} 
In \textbf{Table,\,\ref{tab: example_app}}, we show five randomly selected greedy as well as nucleus sampled (total 10) generations with temperature = $0.8$, top\_p = $0.95$ and the maximum generated tokens set to $50$.

\begin{table}[H]
    \centering
    \caption{\footnotesize{
   \rev{We show Spec-CoFrGeNet generations where, FFNs are replaced with the architecture in Fig. \ref{fig:mlp} in the stated models below.}
    }}
    \vspace{2mm}
    \small 
    \begin{tabular}{p{12.5cm}}
    \toprule   
    \textbf{Example 1} \\    
    \midrule
    \textbf{Prompt:} 'Once upon a time, in a small village,' \\
    \textbf{Greedy:} Once upon a time, in a small village, there lived a young girl named Lily. Lily was known for her kind heart and her love for animals. She had a special bond with a small white dog named Max, who was her constant companion. One sunny morning, as Lily was walking through the\\
    \textbf{Sampled:} Once upon a time, in a small village, there lived a wise old owl named Boosie. Boosie was known for his wisdom and his unique ability to see into the future. One day, as Boosie was on a stroll through the woods, he came upon a beautiful,\\
    \\
    \toprule
    \textbf{Example 2} \\  
    \midrule
    \textbf{Prompt:} 'The theory of relativity states that'\\
    \textbf{Greedy:} The theory of relativity states that the speed of light is constant in all reference frames, and that time and space are relative to the observer's frame of reference. The concept of time dilation and length contraction are fundamental to the theory of relativity, and have been extensively tested and confirmed\\
    \textbf{Sampled:} The theory of relativity states that the speed of light is constant and does not depend on the relative motion of the observer. This has significant implications for the behavior of objects in high-speed collisions. The concept of mass is central to the theory of relativity, and it is a measure\\
    \\
    \toprule
    \textbf{Example 3} \\  
    \midrule
    \textbf{Prompt:} 'Artificial intelligence is'\\
    \textbf{Greedy:} Artificial intelligence is a broad term that refers to the ability of machines to perform tasks that would typically require human intelligence. This includes tasks such as learning, reasoning, and problem-solving. AI is being used in a wide range of applications, from self-driving cars to virtual\\
    \textbf{Sampled:} Artificial intelligence is a form of computer technology that mimics human intelligence. While this is not a new concept, it is not widely known that AI has been used in a variety of ways throughout history, including to predict the weather, write poetry, and solve mathematical problems\\
    \\
    \toprule
    \textbf{Example 4} \\  
    \midrule
    \textbf{Prompt:} 'The most important discovery in biology was'\\
    \textbf{Greedy:} The most important discovery in biology was the discovery of the cell. The cell is the basic unit of life. It is the smallest unit of an organism that can perform all the functions necessary for life. The cell is the basic unit of life. It is the smallest unit of an organism\\
    \textbf{Sampled:} The most important discovery in biology was that the chromosomes of all living organisms, regardless of species, are packed into a tiny nucleus. The chromosomes, and their structure, is fundamentally the same across species. This discovery also demonstrated that DNA is the key component to the genetic code. The discovery\\
    \\
    \toprule
    \textbf{Example 5} \\  
    \midrule
    \textbf{Prompt:} 'In the year 2050, cities will'\\
    \textbf{Greedy:} In the year 2050, cities will be home to 70\% of the world's population. The world's population is expected to reach 9.7 billion by 2050, and the number of people living in cities is expected to increase by 2.5 billion. The\\
    \textbf{Sampled:} In the year 2050, cities will represent 50\% of the global population and 80\% of the GDP, while rural areas will have a population of 40\% and a GDP of 4 trillion. This means that cities will have a significant impact on global economic and social development\\
    \bottomrule
    \end{tabular}
    \label{tab: example_app}
\end{table}

\begin{figure}[ht]
\centering
\includegraphics[width=.32\textwidth]{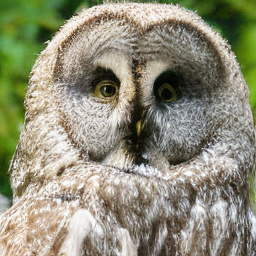}
%\hspace{25mm}
\includegraphics[width=.32\textwidth]{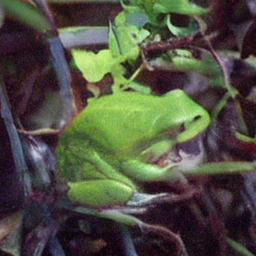}
\includegraphics[width=.32\textwidth]{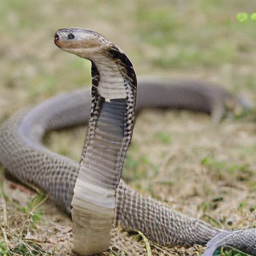}\\
\includegraphics[width=.32\textwidth]{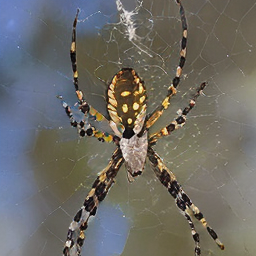}
%\hspace{25mm}
\includegraphics[width=.32\textwidth]{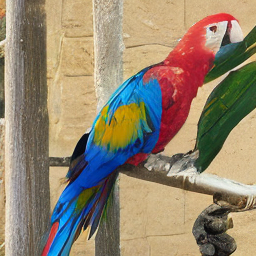}
\includegraphics[width=.32\textwidth]{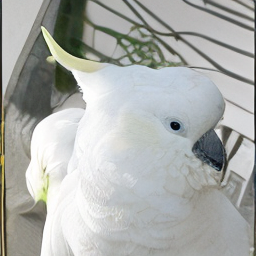}\\
\includegraphics[width=.32\textwidth]{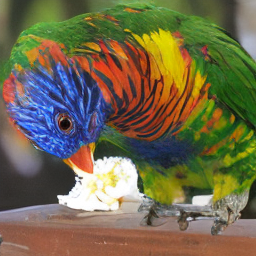}
%\hspace{25mm}
\includegraphics[width=.32\textwidth]{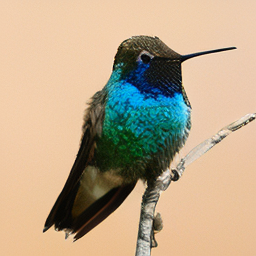}
\includegraphics[width=.32\textwidth]{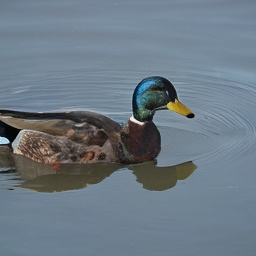}
  \caption{\rev{Generated images from DDT-XL/2 trained using Spec-CoFrGeNet after replacing FFNs, which has $\approx 30$\% fewer parameters than the standard model.}}
\label{fig:diffusionex}
%\vspace{-0.3cm}
\end{figure}

\rev{\paragraph{DDT-XL/2}
In Fig. \ref{fig:diffusionex} we see randomly selected nine example generations of Spec-CoFrGeNet in a diffusion model.}

\section{Ablation results}

Here we present some results ablating the various parameters of \name. 
As seen in \textbf{Table\,\ref{tab: gpt_app}, \ref{tab: granite_app}, \ref{tab: llama_app}}, we present results for models using ladders with depth and width of 1. The performance of models with the higher depth and width reported before are either similar or better. 

\begin{table}[htbp]
    \centering
    \caption{\small{\rev{Perplexity of pretrained Spec-CoFrGeNet replacing FFNs in 
    \textbf{GPT-2 XL} with depth and width $1$.}}}
    \label{tab: gpt_app}
    \setlength{\tabcolsep}{7pt}
    \renewcommand{\arraystretch}{1.15}
    \resizebox{\linewidth}{!}{%
    \begin{tabular}{c|c|c|c|c|c|c|c|c}
        \toprule
        \toprule
        \textbf{Dataset} &
        \textbf{Method} & 
        % \textbf{\shortstack{Expansion\\ratio}} 
        % \textbf{$\alpha=1$} 
        % &
        \textbf{Params} &
        \textbf{PTB} &
        \textbf{Wikitext2} &
        \textbf{Lambada} &
        \textbf{AgNews} &
        \textbf{LM1B} &
        \textbf{Wikitext103} \\
        \midrule

        OWT
        & \cellcolor{lightblue} \name
        & \cellcolor{lightblue} 985M
        & \cellcolor{lightblue} 29.89
        & \cellcolor{lightblue} 17.16
        & \cellcolor{lightblue}  8.13
        & \cellcolor{lightblue}  35.65
        & \cellcolor{lightblue} 40.21
        & \cellcolor{lightblue} 16.17 \\

        \midrule

        GW
        & \cellcolor{lightblue} \name
        & \cellcolor{lightblue}985M
        & \cellcolor{lightblue} 29.72
        & \cellcolor{lightblue}  18.09
        & \cellcolor{lightblue} 30.59
        & \cellcolor{lightblue}  41.63
        & \cellcolor{lightblue} 46.86
        & \cellcolor{lightblue}  18.07
        \\

        \bottomrule
    \end{tabular}
    }
\end{table}

\begin{table}[H]
    \centering
    \caption{\small{\rev{Zero-shot accuracies on open domain Q\&A, reasoning and text understanding tasks for Spec-CoFrGeNet replacing FFNs in \textbf{Granite 4.1} with depth and width $1$.}}}
    \vspace{0.4pt}
    \label{tab: granite_app}
    \setlength{\tabcolsep}{7pt}
    \renewcommand{\arraystretch}{1.15}
    \resizebox{\linewidth}{!}{%
    \begin{tabular}{c|c|c|c|c|c|c|c|c|c|c}
        \toprule
        \toprule
        \textbf{Method} & \textbf{Setting} &

        \textbf{Params} &
        \textbf{OpenBookQA} &
        \textbf{PIQA} &
        \textbf{ARC-Easy} &
        \textbf{WinoGrande} &
        \textbf{HellaSwag} &
        \textbf{LAMBADA} &
        \textbf{BoolQ} &
        \textbf{SciQ}\\
        \midrule

        Spec-CoFrGeNet & FG  & 2.14B& \cellcolor{lightblue}36.8
& \cellcolor{lightblue}73.9
& \cellcolor{lightblue}69.5
& \cellcolor{lightblue}57.8
& \cellcolor{lightblue}59
& \cellcolor{lightblue}49.2
& \cellcolor{lightblue}62.7
& \cellcolor{lightblue}90.7
 \\
    
        \cline{2-11}
        \noalign{\vskip 1.5pt}

        Spec-CoFrGeNet & FNG & 1.1B& \cellcolor{lightblue}35.8
& \cellcolor{lightblue}73
& \cellcolor{lightblue}65.4
& \cellcolor{lightblue}55.6
& \cellcolor{lightblue}53
& \cellcolor{lightblue}45.3
& \cellcolor{lightblue}59.6
& \cellcolor{lightblue}88.3
\\

        \bottomrule
    \end{tabular}
    }
\end{table}

\begin{table}[H]
    \centering
    \caption{\small{\rev{Zero-shot accuracies on open domain Q\&A, reasoning and text understanding tasks for Spec-CoFrGeNet replacing FFNs in \textbf{Llama-MoE} with depth and width 1.}}}
    \vspace{0.2pt}
    \label{tab: llama_app}
    \setlength{\tabcolsep}{7pt}
    \renewcommand{\arraystretch}{1.15}
    \resizebox{\linewidth}{!}{%
    \begin{tabular}{c|c|c|c|c|c|c|c|c|c}
        \toprule
        \toprule
        \textbf{Model} & 
        % \textbf{\shortstack{Expansion\\ratio}} 
        % \textbf{$\alpha=1$} 
        % &
        \textbf{Params} &
        \textbf{OpenBookQA} &
        \textbf{PIQA} &
        \textbf{ARC-Easy} &
        \textbf{WinoGrande} &
        \textbf{HellaSwag} &
        \textbf{LAMBADA} &
        \textbf{BoolQ} &
        \textbf{SciQ}\\
        \midrule

       \cellcolor{lightblue} \name
        & \cellcolor{lightblue} 204M
        & \cellcolor{lightblue}  15.2
        & \cellcolor{lightblue}  61.2
        & \cellcolor{lightblue}  51.2 
        & \cellcolor{lightblue}  50 
        & \cellcolor{lightblue} 28.1 
        & \cellcolor{lightblue}  21
        & \cellcolor{lightblue}  53.4
        & \cellcolor{lightblue}  77.7
        \\

        \bottomrule
    \end{tabular}
    }
\end{table}

We performed experiments with the PL reparametrization on Granite 4.1 for language modeling. As seen in \textbf{Table\,\ref{tab: granite_app_pl}}, we present results for a model trained with ladders of depth and width 5. This produces performance similar to the SPD reparametrization.

\begin{table}[H]
    \centering
    \caption{\small{\rev{Zero-shot accuracies on open domain Q\&A, reasoning and text understanding tasks for Spec-CoFrGeNet replacing FFNs in \textbf{Granite 4.1} with depth and width $5$ using the \textbf{PL} reparametrization.}}}
    \vspace{0.4pt}
    \label{tab: granite_app_pl}
    \setlength{\tabcolsep}{7pt}
    \renewcommand{\arraystretch}{1.15}
    \resizebox{\linewidth}{!}{%
    \begin{tabular}{c|c|c|c|c|c|c|c|c|c|c}
        \toprule
        \toprule
        \textbf{Method} & \textbf{Setting} &

        \textbf{Params} &
        \textbf{OpenBookQA} &
        \textbf{PIQA} &
        \textbf{ARC-Easy} &
        \textbf{WinoGrande} &
        \textbf{HellaSwag} &
        \textbf{LAMBADA} &
        \textbf{BoolQ} &
        \textbf{SciQ}\\
        \midrule

        Spec-CoFrGeNet & FG  & 2.14B& \cellcolor{lightblue}36.9
& \cellcolor{lightblue}73.9
& \cellcolor{lightblue}69.9
& \cellcolor{lightblue}58.5
& \cellcolor{lightblue}59.1
& \cellcolor{lightblue}48.7
& \cellcolor{lightblue}62.6
& \cellcolor{lightblue}90.8
 \\
    
        \cline{2-11}
        \noalign{\vskip 1.5pt}

        Spec-CoFrGeNet & FNG & 1.1B& \cellcolor{lightblue}35.4
& \cellcolor{lightblue}72.9
& \cellcolor{lightblue}63.9
& \cellcolor{lightblue}55.7
& \cellcolor{lightblue}53.3
& \cellcolor{lightblue}45.1
& \cellcolor{lightblue}59.4
& \cellcolor{lightblue}86.3
\\

        \bottomrule
    \end{tabular}
    }
\end{table}

In \name, $\delta$ is not a stabilizing hyperparameter unlike $\epsilon$ in CoFrGeNet. It is simply an offset so that the zero eigenvalue do not lie at the edge of the Gershgorin disk (see Section 4) and can thus be arbitrarily small.  
In our experiments, we choose $\delta = 1e-3$, which is extremely small and it leads to desirable performance as demonstrated in the paper. Thus, we do not need to try higher values.

Nonetheless to empirically verify our reasoning, we trained a DDT-XL/2 model with a higher $\delta$
 as shown in \textbf{Table\,\ref{tab:image_generation_app}}. We observe that the results for the higher 
 are only slightly worse in FID as expected, where training was stable. Thus \name is robust and does not need tuning of this offset, saving expensive pretraining runs.

\begin{table}[H]
    \centering
    \caption{\rev{Performance of various metrics using DDT-XL/2 for \name at different $\delta$s. $\uparrow$ refers to higher the better and $\downarrow$ refers to lower the better.}}
    \label{tab:image_generation_app}
    \setlength{\tabcolsep}{7pt}
    \renewcommand{\arraystretch}{1.15}
    \resizebox{\textwidth}{!}{%
    \begin{tabular}{c|c|c|c|c|c|c}
        \toprule
        \textbf{Method} &
        \textbf{$\delta$} &
        \textbf{FID $\downarrow$} &
        \textbf{sFID $\downarrow$} &
        \textbf{Inception Score $\uparrow$} &
        \textbf{Precision $\uparrow$} &
        \textbf{Recall $\uparrow$} \\
        \midrule

        \multirow{2}{*}{\name} &
        \cellcolor{lightblue} $10^{-3}$ &
        \cellcolor{lightblue} 1.606 &
        \cellcolor{lightblue} 4.406 &
        \cellcolor{lightblue} 290.835 &
        \cellcolor{lightblue} 0.621 &
        \cellcolor{lightblue} 0.799 \\

         &
         $10^{-2}$ &
       1.612 &
      4.475 &
        292.564 &
       0.621 &
     0.799\\

        \bottomrule
    \end{tabular}
    }
\end{table}
% \vspace{-2mm}

\end{document}